\documentclass[letterpaper, 10 pt, conference]{ieeeconf}  

\IEEEoverridecommandlockouts                              

\usepackage{amsmath} 
\usepackage{amssymb}
\usepackage{graphicx} 
\usepackage{subcaption}
\usepackage{multirow}
\usepackage{booktabs}
\usepackage{balance}

\title{\LARGE \bf
G-EDF-Loc: 3D Continuous Gaussian Distance Field for Robust Gradient-Based 6DoF Localization
}

\author{José E. Maese$^{1}$, Lucía Coto-Elena$^{1}$, Luis Merino$^{1}$ and Fernando Caballero$^{1}$
\thanks{*This work was supported by the grants PICRA 4.0 (PLEC2023-010353), funded by the Spanish Ministry of Science and Innovation and the Spanish Research Agency (MCIN/AEI/10.13039/501100011033); and COBUILD (PID2024-161069OB-C31), funded by the Spanish Ministry of Science, Innovation and Universities, the Spanish Research Agency (MICIU/AEI/10.13039/501100011033) and the European Regional Development Fund (FEDER, UE).}
\thanks{$^{1}$The authors are with the Service Robotics Laboratory, Universidad Pablo de Olavide, Seville, Spain. {\tt\small \{jemaealv,lcotele,lmercab,fcaballero\}@upo.es}}%
}

\begin{document}

\maketitle
\thispagestyle{empty}
\pagestyle{empty}

\begin{abstract}

This paper presents a robust 6-DoF localization framework based on a direct, CPU-based scan-to-map registration pipeline. The system leverages G-EDF, a novel continuous and memory-efficient 3D distance field representation. The approach models the Euclidean Distance Field (EDF) using a Block-Sparse Gaussian Mixture Model with adaptive spatial partitioning, ensuring $C^1$ continuity across block transitions and mitigating boundary artifacts. By leveraging the analytical gradients of this continuous map, which maintain Eikonal consistency, the proposed method achieves high-fidelity spatial reconstruction and real-time localization. Experimental results on large-scale datasets demonstrate that G-EDF-Loc performs competitively against state-of-the-art methods, exhibiting exceptional resilience even under severe odometry degradation or in the complete absence of IMU priors.


\end{abstract}

\section{Introduction}
\label{sec:introduction}

Precise 6-DoF localization is a fundamental pillar of robust autonomous navigation, requiring systems to efficiently align real-time sensor data with pre-built environmental models. While modern 3D LiDAR sensors provide dense, long-range geometric measurements, directly processing these massive, unstructured point clouds for real-time state estimation introduces significant computational overhead. In general, the efficiency of standard registration algorithms is intrinsically tied to the quantity of points the model works with during each alignment step. As a result, these approaches can suffer from performance bottlenecks or reduced stability under aggressive motions. To achieve smoother and more resilient pose tracking, there is a fundamental advantage in using map representations that explicitly model the Euclidean Distance Field (EDF). This enables direct gradient-based optimization, avoiding the matching ambiguities of raw point cloud registration and ensuring a robust convergence regardless of the initial odometry state.

Current mapping paradigms frequently impose a trade-off between computational efficiency and geometric continuity. Feature-based strategies \cite{LOAM, FASTLIO2} achieve real-time performance by extracting sparse geometric primitives, yet the resulting models lack the continuous volumetric context necessary for stable gradient-based alignment. Dense grid-based structures, such as TSDFs or ESDFs \cite{Voxblox, FIESTA}, enable fast distance queries but incur substantial memory overhead in large-scale environments. Furthermore, their reliance on trilinear interpolation yields only $C^0$ continuity, producing discrete gradient transitions at voxel boundaries that might hinder the convergence of non-linear optimization frameworks. While recent Implicit Neural Representations (INRs) \cite{iSDF, SHINE_Mapping} resolve these artifacts by providing continuous field approximations, their extensive computational and memory demands typically necessitate dedicated GPU hardware, limiting their applicability on resource-constrained platforms.

\begin{figure}[t!]
    \centering
    \includegraphics[width=0.49\columnwidth]{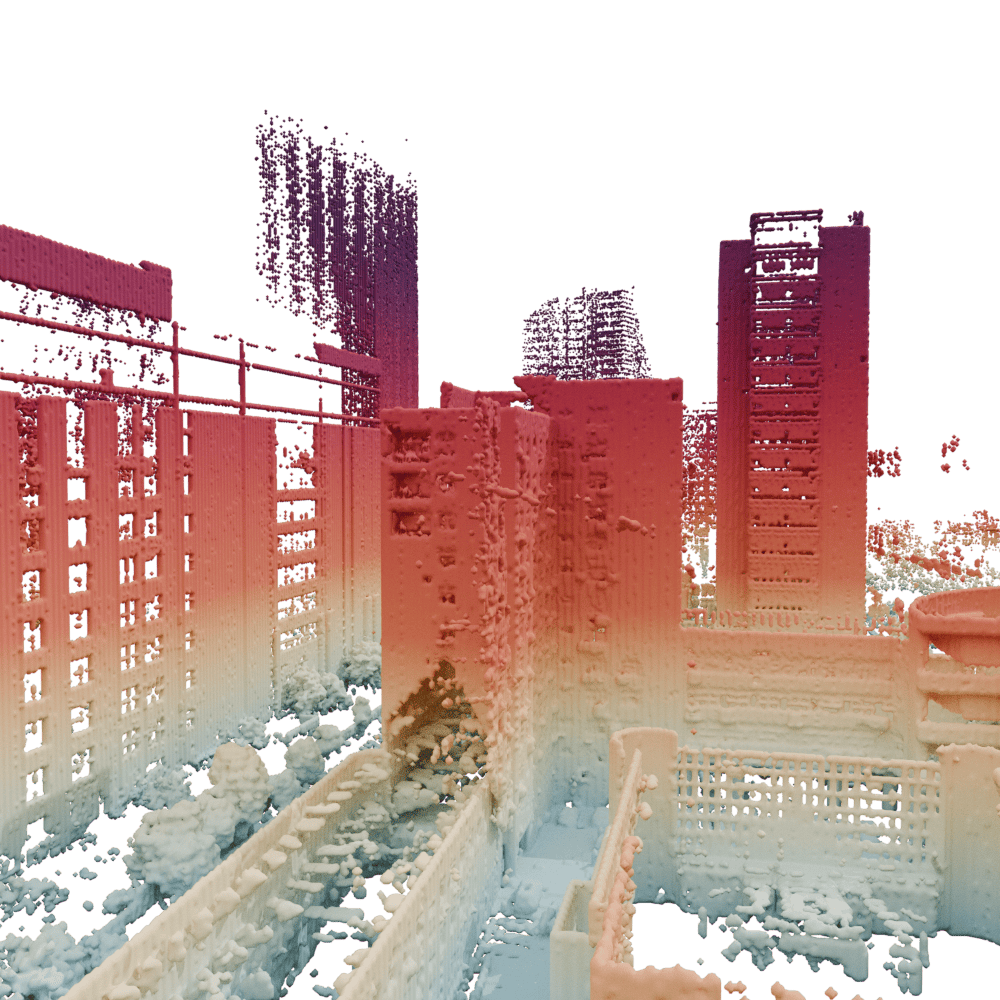}
    \includegraphics[width=0.49\columnwidth]{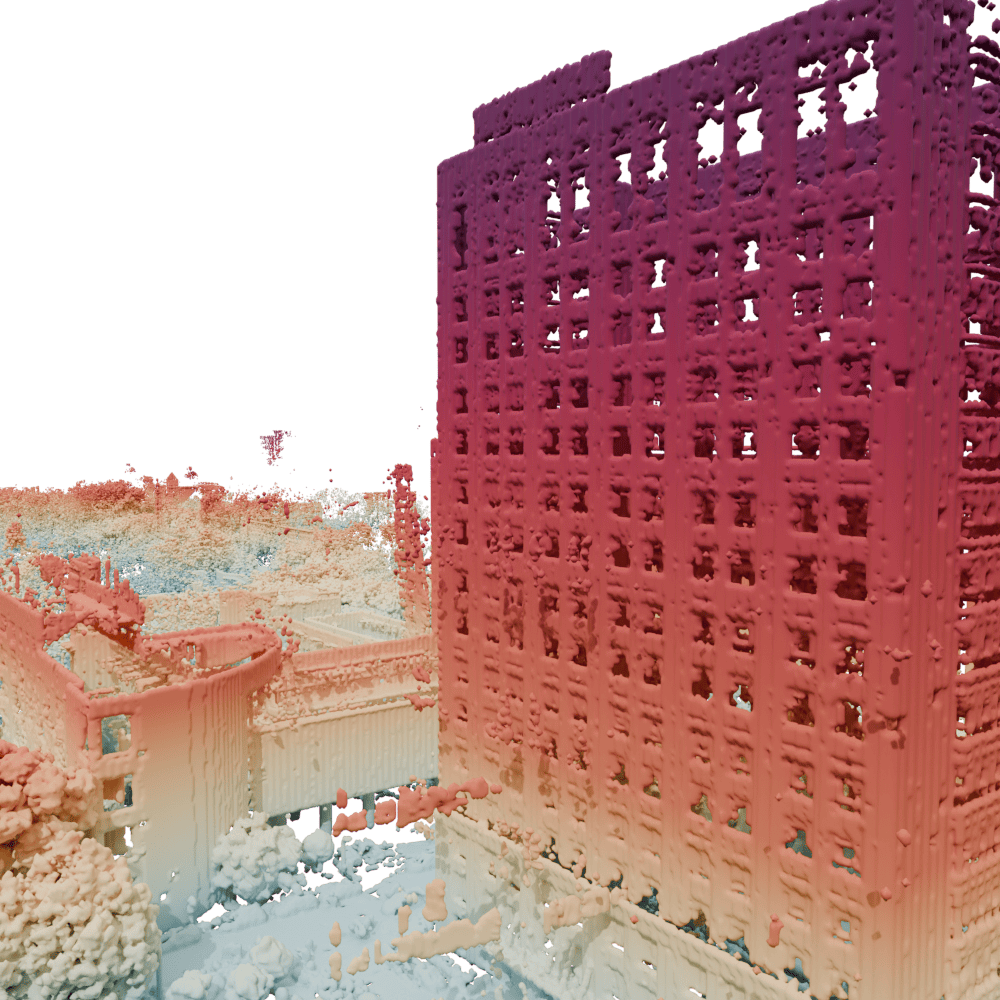}
    \includegraphics[width=\columnwidth]{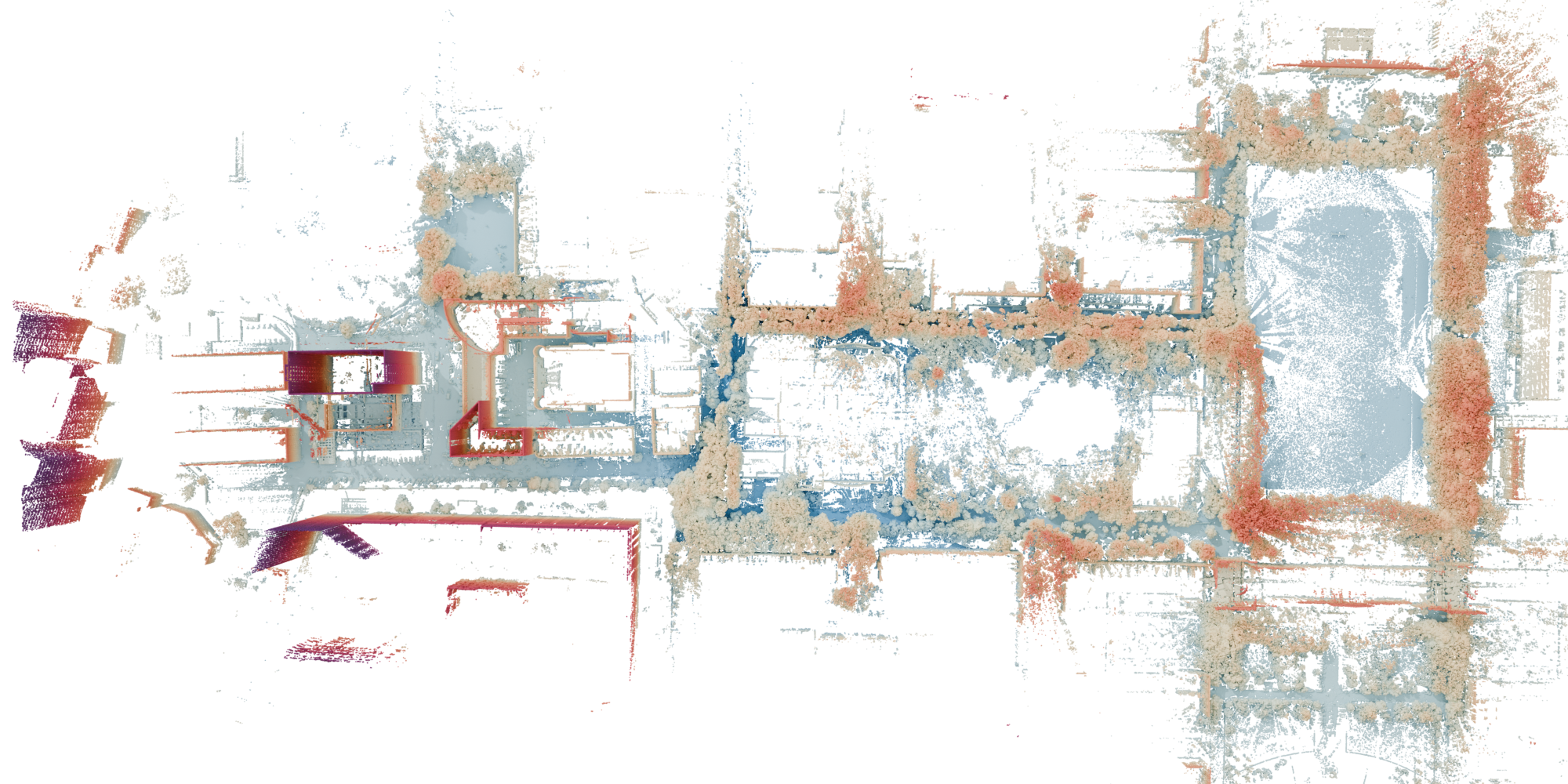}
    
    \caption{Overview of the Snail dataset environment reconstruction. Top: detailed perspectives of the large-scale outdoor structures and local elevation changes. Bottom: a top-down view showing the full $900 \times 610 \times 120$ m extent of the environment.}
    \label{fig:snail_overview}
\end{figure}

To address these limitations, this work introduces G-EDF-Loc, a robust 6-DoF localization system built upon a novel continuous distance field representation. In contrast to traditional probabilistic occupancy frameworks, G-EDF leverages anisotropic Gaussians as universal geometric function approximators to explicitly regress the distance field. By substituting rigid discrete voxels with adaptively blended spatial blocks, the proposed architecture achieves exceptional compression rates while strictly guaranteeing global $C^1$ continuity. As illustrated in Fig.~\ref{fig:snail_overview}, this formulation enables the scalable representation of unbounded environments. A primary advantage of G-EDF is the derivation of closed-form analytical gradients that empirically approximate the Eikonal equation ($\|\nabla \hat{d}\| \approx 1$). This mathematical property facilitates highly efficient, CPU-based inference, establishing a rigorous foundation for precise gradient-based localization and trajectory optimization.

\section{Related Work}
\label{sec:related_work}

Reliable autonomous navigation in large-scale environments relies fundamentally on map-based localization, where the robot estimates its pose by aligning sensor data against a pre-built global representation.

\subsection{Map Representations}

To facilitate localization and planning, maps are typically discretized into volumetric grids. While OctoMap \cite{octomap} is the standard for occupancy, gradient-based techniques rely on explicit distance information. Frameworks like Voxblox \cite{Voxblox} and FIESTA \cite{FIESTA} enable efficient CPU-based computation of Euclidean Signed Distance Fields (ESDFs). Sparse data structures, such as OpenVDB \cite{OpenVDB}, have been adopted to improve scalability in large environments \cite{VDBFusion}. Recent advancements like DB-TSDF \cite{dbtsdf} further enhance these representations by incorporating directionality to improve reconstruction fidelity. However, regardless of the data structure, these grid-based methods rely on trilinear interpolation to query values between voxels. This yields only $C^0$ continuity, resulting in discontinuous gradients at voxel boundaries. These discontinuities can introduce local minima and hinder the convergence of non-linear optimizers used in localization tasks such as DLL \cite{iros21dll}.

To overcome discretization artifacts, research has pivoted toward continuous representations. Implicit Neural Representations (INRs) approximate the field using neural networks (MLPs). Methods like iSDF \cite{iSDF} and SHINE-Mapping \cite{SHINE_Mapping} allow for resolution-independent queries with smooth gradients. Nevertheless, INRs typically demand significant GPU resources for training and inference, posing challenges for resource-constrained aerial or mobile robots. Alternatively, Gaussian Process (GP) approaches, such as VDB-GPDF \cite{VDB-GPDF}, provide mathematically continuous fields with uncertainty quantification. However, GP inference typically scales cubically ($O(N^3)$), necessitating approximations that complicate real-time deployment.

Kernel-based and Gaussian representations offer a computationally tractable middle ground. Classic kernel-based approaches like Hilbert Maps \cite{HilbertMaps} utilized kernels to model continuous occupancy, but lacked explicit distance information. While NDT \cite{1249285} models local surface probabilities by fitting a single Gaussian distribution within each voxel, recent computer vision approaches like 3D Gaussian Splatting (3DGS) \cite{3dgs} use anisotropic Gaussians for radiance field rendering. However, 3DGS minimizes photometric error rather than geometric error, producing artifacts unsuitable for robotic collision checking \cite{SuGaR}. In contrast, our proposed G-EDF representation employs anisotropic Gaussians as geometric function approximators to explicitly regress the Euclidean Distance Field. This approach combines the computational efficiency of analytical functions (running on CPU) with the rigorous $C^1$ continuity required for stable gradient-based localization.

\subsection{Point Cloud Registration and Localization}

Probabilistic approaches, such as Monte Carlo Localization (MCL) \cite{Perezgrau17JARS} and NDT-MCL \cite{6696380}, handle sensor noise and global uncertainty using particle filters. While robust, they are computationally intensive and their accuracy depends heavily on the number of particles. Alternatively, optimization-based methods seek to minimize an alignment error.

The Iterative Closest Point (ICP) algorithm \cite{121791} remains a baseline but suffers from high computational costs due to nearest-neighbor searches. Highly optimized, multi-threaded variants such as Fast-GICP, frequently integrated into state-of-the-art frameworks like \textit{hdl\_localization} \cite{koide2019portable}, significantly mitigate this issue. However, their computational efficiency remains intrinsically bounded by the quantity of points the model works with during each optimization step. To avoid exhaustive point matching, the Normal Distributions Transform (NDT) discretizes the map into cells modeled by Gaussian distributions, optimizing a probability density function. Modern implementations of both Fast-GICP and NDT within the \textit{hdl\_localization} package establish a rigorous and robust baseline for discrete, CPU-based point cloud registration.

More recently, gradient-based methods utilizing Distance Fields, such as DLL \cite{iros21dll}, have demonstrated superior efficiency by treating localization as a direct optimization problem over a distance field. However, the performance of these optimizers is intrinsically bounded by the quality and continuity of the underlying map representation. 

Our proposed G-EDF-Loc framework addresses these fundamental limitations by leveraging a $C^{1}$ continuous Gaussian distance field. Unlike traditional methods that suffer from severe computational bottlenecks or failure to converge under aggressive motions or missing IMU data, G-EDF provides stable analytical gradients that ensure robust convergence even in the absence of initial odometry priors. As demonstrated in our experimental results, while state-of-the-art discrete baselines such as Fast-GICP and NDT exhibit unstable pose estimates and extreme spikes in processing time under high-noise conditions, our approach maintains a robust performance. By substituting rigid discrete voxels with adaptively blended spatial blocks, we provide a mathematically consistent field that facilitates global localization resilience where traditional discrete representations fail.

\section{Continuous Distance Field Representation}
\label{sec:method}

This section details the proposed framework, designed to represent large-scale 3D environments as a continuous Block-Sparse Gaussian Distance Field. Unlike traditional voxel grids or octrees, which discretize space, the proposed approach models the Euclidean Distance Field (EDF) as a continuous function approximated by a mixture of anisotropic Gaussians.

\subsection{Mathematical Formulation}

The core concept is to employ a Gaussian Mixture Model (GMM) not as a probability density function, but as a universal function approximator. Let $d_{GT}(\mathbf{x}) \ge 0$ be the ground truth distance value at a query point $\mathbf{x} \in \mathbb{R}^3$, representing the distance to the nearest surface. This field $\hat{d}(\mathbf{x})$ is approximated as a weighted sum of $K$ anisotropic Gaussian kernels:
\begin{equation}
    \hat{d}(\mathbf{x}) = \sum_{k=1}^{K} w_k \cdot \exp\left( -\frac{1}{2} (\mathbf{x} - \boldsymbol{\mu}_k)^T \mathbf{\Sigma}_k^{-1} (\mathbf{x} - \boldsymbol{\mu}_k) \right)
\end{equation}

\noindent where, for each $k$-th Gaussian, $w_k \in \mathbb{R}$ is the amplitude (weight), $\boldsymbol{\mu}_k \in \mathbb{R}^3$ is the mean (center), and $\mathbf{\Sigma}_k$ is the covariance matrix.

To maximize inference speed, the covariance matrix is constrained to be axis-aligned. Although this limits individual rotational expressiveness, Gaussian mixtures remain universal function approximators \cite{BRFN}. Our adaptive complexity strategy (Sec. III-C.2) compensates for this by dynamically increasing the kernel density $K$ along arbitrarily slanted surfaces. 

By bypassing the computational bottleneck of evaluating full covariance matrices with local rotations, the model achieves a highly efficient parameterization. Specifically, the inverse covariance is defined using a vector of length scales $\mathbf{l}_k = [l_{kx}, l_{ky}, l_{kz}]^T$, yielding the diagonal form $\mathbf{\Sigma}_k^{-1} = \text{diag}(l_{kx}^{-2}, l_{ky}^{-2}, l_{kz}^{-2})$. The expression for a single Gaussian kernel $g_k(\mathbf{x})$ simplifies to:
\begin{equation}
    g_k(\mathbf{x}) = w_k \cdot \exp\left( -\frac{1}{2} \sum_{j \in \{x,y,z\}} \frac{(x_j - \mu_{kj})^2}{l_{kj}^2} \right).
\end{equation}

Notably, unlike probabilistic GMMs, the weights $w_k$ are not constrained to sum to one and can be negative. While the target distance field is non-negative, negative weights play a crucial role in the approximation. They act as subtractive terms that enable the model to carve out sharp "valleys" (minima) in the field. This is particularly important in regions with high point density, where the superposition of multiple positive Gaussians would otherwise overestimate the distance. Negative components facilitate the convergence to zero at the surface manifold, ensuring a tight fit.

\subsection{Optimization and Training}

The parameters $\theta = \{ w_k, \boldsymbol{\mu}_k, \mathbf{l}_k \}_{k=1}^K$ are estimated by minimizing the squared residual between the approximation and the local Euclidean Distance Transform (EDT). The optimization problem over a set of sample points $\mathcal{S}$ is defined as:
\begin{equation}
    \min_{\theta} \sum_{\mathbf{x} \in \mathcal{S}} \left( \hat{d}(\mathbf{x}) - d_{GT}(\mathbf{x}) \right)^2.
\end{equation}

The non-linear least squares problem is solved using the Levenberg-Marquardt algorithm implemented in the Ceres Solver framework \cite{Agarwal_Ceres_Solver_2022}. Analytical derivatives for all parameters are computed to accelerate convergence.

\subsection{Block-Sparse Architecture}

To ensure scalability in unbounded environments, a streaming, block-sparse architecture is employed. The 3D space is spatially hashed into a grid of independent cubes. This structure allows handling large-scale point clouds by processing local regions in parallel.

\subsubsection{Local Ground Truth Generation}
The system processes not only cubes containing surface measurements but also empty cubes whose centers lie within a specified distance of any point. This ensures that the immediate free space surrounding the surface is accurately modeled, maintaining field continuity. For each active cube, a local high-resolution voxel grid is constructed and its exact Euclidean Distance Transform (EDT) is computed. This dense local field serves as the high-fidelity ground truth for the optimization process. By training against the local EDT rather than raw points, the optimization is decoupled from the point density of the sensor.

\subsubsection{Adaptive Complexity}
A defining feature of the framework is the adaptive resource allocation strategy, which automatically determines the necessary model complexity for each block. Rather than imposing a fixed number of kernels $K$, a progressive fitting approach is employed. The optimization is initialized by performing Non-Maximum Suppression (NMS) on the local EDT grid to identify local extrema. Specifically, voxels that are strictly greater (maxima) or smaller (minima) than their spatial neighbors are identified. Positive Gaussian centroids $\boldsymbol{\mu}_k$ are initialized at these local maxima (representing the medial axis of free space), while negative centroids are placed at local minima (representing the surface or object boundaries). This initialization provides a warm-start that closely approximates the topological skeleton of the distance field, significantly reducing the risk of getting trapped in poor local minima during the non-convex optimization.

The training begins with a minimal set of kernels, particularly for the neighboring empty cubes where the EDT geometry is typically simpler and easier to approximate. Upon convergence, the Mean Absolute Error (MAE) is evaluated; if the reconstruction error exceeds a tolerance threshold, the capacity $K$ is increased, and the model is refined. This iterative process ensures that geometrically simple regions are represented with minimal memory footprint, while complex structures receive the higher kernel density required to preserve fine details.

\begin{figure}[t!]
    \centering
    \begin{subfigure}[b]{0.45\columnwidth}
        \centering
        \raggedleft 
        \includegraphics[height=3.2cm, keepaspectratio]{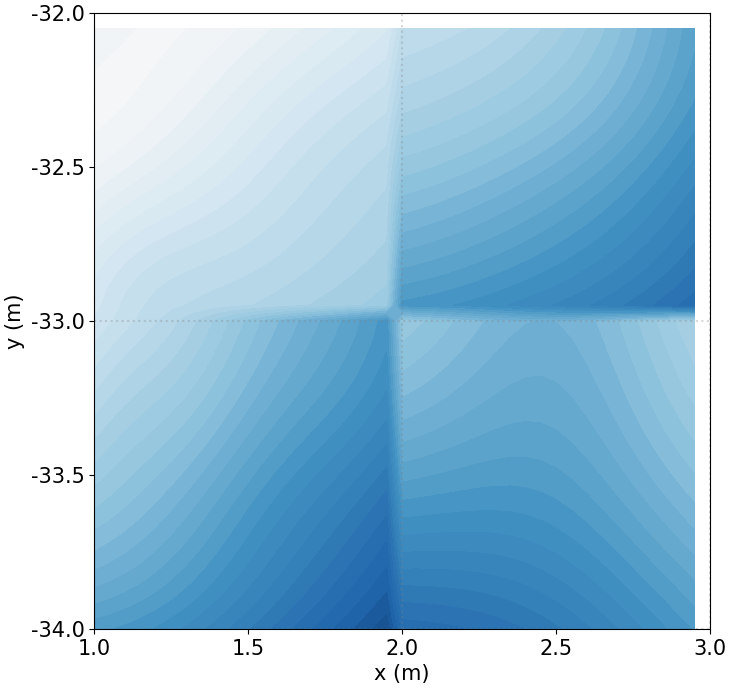}
        \caption{EDF (Blend False)}
        \label{fig:edf_false}
    \end{subfigure}
    \hfill 
    \begin{subfigure}[b]{0.53\columnwidth}
        \centering
        \raggedright
        \includegraphics[height=3.2cm, keepaspectratio]{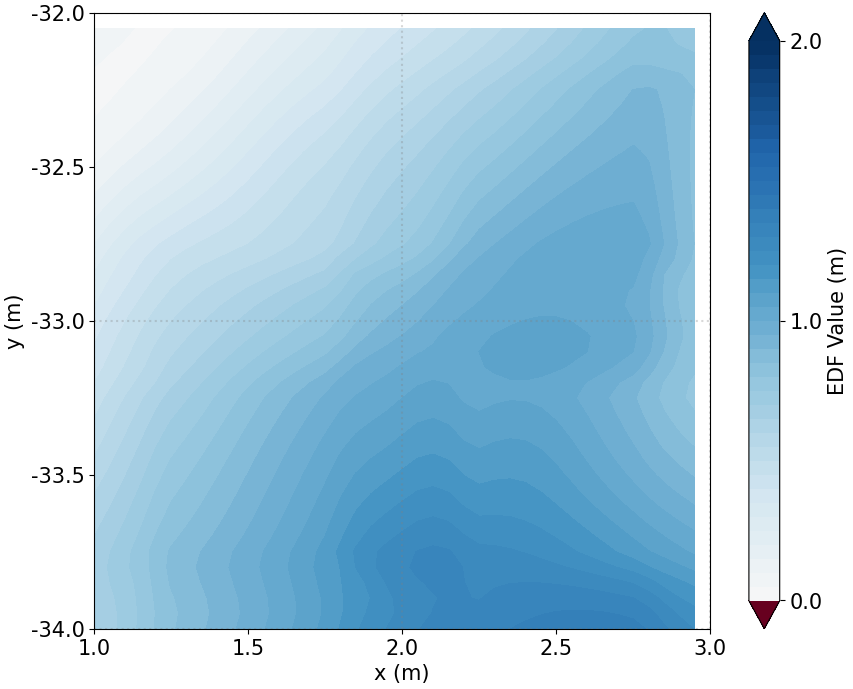}
        \caption{EDF (Blend True)}
        \label{fig:edf_true}
    \end{subfigure}
    
    \vspace{0.2cm} 
    
    \begin{subfigure}[b]{0.45\columnwidth}
        \raggedleft
        \includegraphics[height=3.2cm, keepaspectratio]{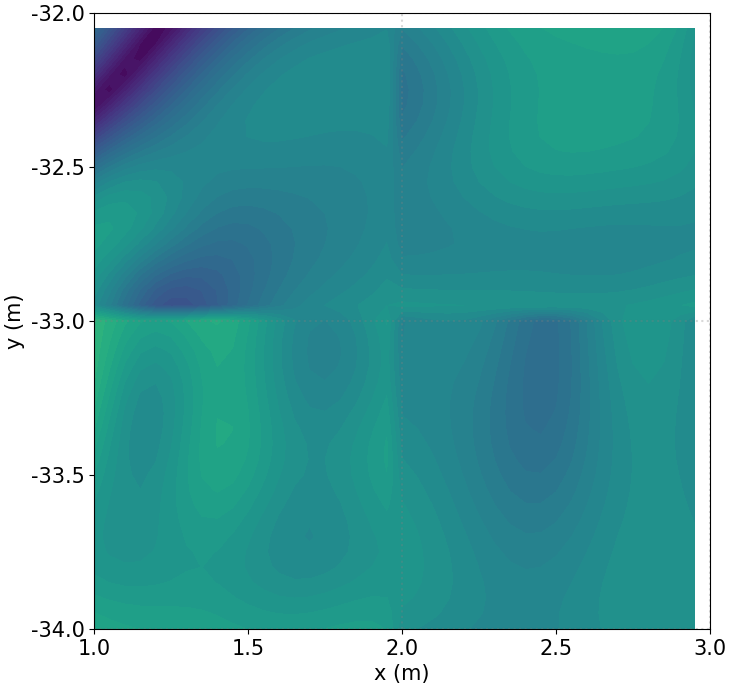}
        \caption{Gradient (Blend False)}
        \label{fig:grad_false}
    \end{subfigure}
    \hfill
    \begin{subfigure}[b]{0.53\columnwidth}
        \raggedright
        \includegraphics[height=3.2cm, keepaspectratio]{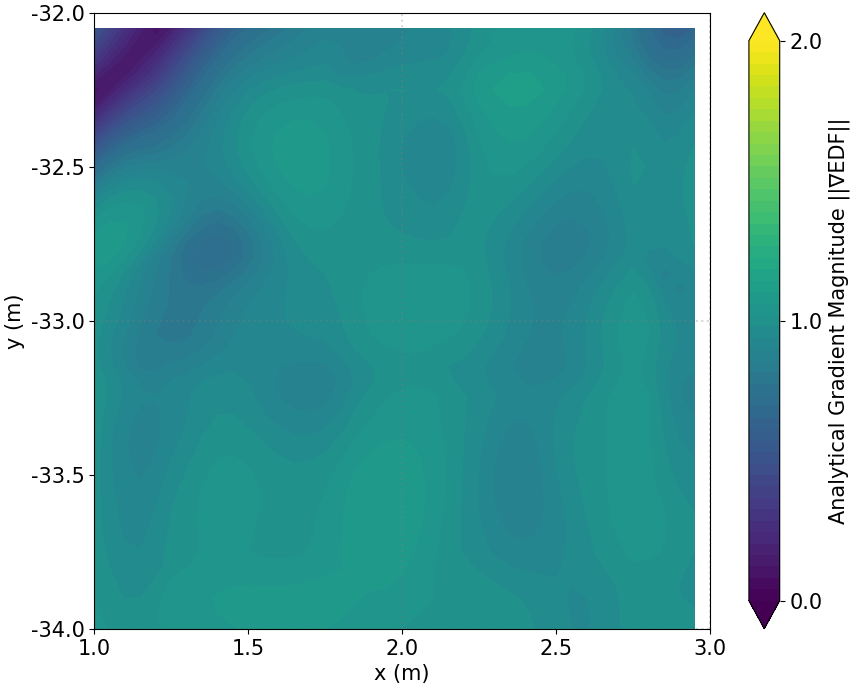}
        \caption{Gradient (Blend True)}
        \label{fig:grad_true}
    \end{subfigure}
    
    \caption{Cross-section analysis on a local subset of the New College dataset ($z=3.0$ m). Parameters: $1.0$ m$^3$ blocks with $\delta=0.25$ m overlap. Left column (a, c): Discontinuities in the distance field and gradient without blending. Right column (b, d): The proposed G-EDF representation, ensuring global $C^1$ continuity.}
    \label{fig:comparacion_college}
\end{figure}

\subsubsection{Reconstruction and Continuity}

Since the optimization is performed locally per cube, discontinuities may arise at boundaries (see Fig. \ref{fig:edf_false}). To enforce global $C^1$ continuity, adjacent blocks share an overlap margin $\delta$. In these transition regions, the final distance value is computed as a weighted blend of the overlapping local fields. The blending weights are derived using the cubic Hermite interpolation polynomial, commonly referred to as the Smoothstep function \cite{ebert2002texturing}:

\begin{equation}
    \alpha(t) = 3t^2 - 2t^3, \quad t \in [0,1],
\end{equation}

\noindent where $t$ represents the normalized position within the margin. Since $\alpha'(0)=\alpha'(1)=0$, this weighting ensures that the gradient transitions smoothly between independent local optimizations without introducing artifacts.

A major advantage of using a Gaussian basis is the existence of a closed-form analytical gradient, which is computationally efficient and essential for trajectory optimization algorithms. The gradient of the field $\nabla \hat{d}(\mathbf{x})$ is derived directly from Eq. (1) and (2) as:
\begin{equation}
    \nabla \hat{d}(\mathbf{x}) = \sum_{k=1}^{K} \nabla g_k(\mathbf{x}) = - \sum_{k=1}^{K} g_k(\mathbf{x}) \cdot \mathbf{\Sigma}_k^{-1} (\mathbf{x} - \boldsymbol{\mu}_k).
\end{equation}

This allows for exact evaluation without numerical differentiation errors.

The blending strategy results in a globally $C^1$ continuous field (Fig. \ref{fig:edf_true}). Furthermore, as shown in Fig. \ref{fig:grad_true} and quantitatively verified in Table \ref{tab:metrics}, the gradient magnitude naturally approximates unity ($\|\nabla \hat{d}\| \approx 1$) throughout the modeled volume. While our Gaussian formulation does not analytically enforce the Eikonal equation by design, minimizing the regression error against the exact local EDT empirically drives the continuous field to inherit this geometric property. This provides consistent and reliable gradient information for localization and navigation tasks, both near boundaries and in open spaces.

\subsection{Map Serialization and Implementation}

To ensure scalability and interoperability, the resulting continuous field is serialized into a compact binary format (`.bin`). The file structure consists of a lightweight header encoding global metadata (spatial bounds, block size, overlap margin $\delta$, and the global mean reconstruction error), followed by the streaming block payload, where each cube encapsulates its specific Gaussian parameters and local error metrics. This self-contained format allows for offline reconstruction and reuse without external dependencies.

To facilitate reproducibility, the code used in this work is publicly available here\footnote{https://github.com/robotics-upo/G-EDF}.

\section{Direct Gradient-Based Localization}
\label{sec:localization}

This section presents G-EDF-Loc, a full 6-DoF localization system that leverages the proposed G-EDF representation. By leveraging the continuous and memory-efficient properties of the Block-Sparse Gaussian Mixture Model, the system is capable of performing fast and accurate gradient-based alignment, scaling seamlessly to large-scale environments \cite{zhang2021multicamera}\cite{huai2025snailradarlargescalediverse}. The localization pipeline requires the pre-computed G-EDF map and an initial pose guess.

The core architecture is driven by an Error-State Kalman Filter (ESKF) that continuously predicts the 6-DoF pose, velocities, and sensor biases of the vehicle (detailed in Section \ref{sec:eskf}). Upon the reception of a new LiDAR scan, the system first utilizes the high-frequency state history from the ESKF to deskew the incoming cloud, compensating for the sensor's continuous motion during the sweep. Then, using the ESKF prediction as an initial guess, the system aligns the deskewed cloud against the continuous G-EDF representation via a non-linear optimization process (detailed in Section \ref{sec:optimization}). This optimization yields a highly accurate pose correction that subsequently updates the filter state.

While the integration of high-frequency IMU measurements provides the optimal prior to bound the optimization search space, it is also crucial for performing the aforementioned deskewing, effectively correcting severe point cloud distortions caused by abrupt or aggressive movements. Nevertheless, the framework is designed to be highly resilient. As will be demonstrated in the experimental study (Section \ref{sec:localization_evaluation}), the system is capable of maintaining accurate localization even in the complete absence of IMU data, relying exclusively on the sequential scan-to-map alignment of successive point clouds.
To provide further implementation details and facilitate reproducibility, the complete source code of the proposed framework is publicly available here\footnote{https://github.com/robotics-upo/G-EDF-Loc}.

\subsection{Prior Guess - Error State Kalman Filter}
\label{sec:eskf}

The filter maintains a 15-Degree-of-Freedom (DoF) state vector $\mathbf{x}$, defined as:
\begin{equation}
\mathbf{x} = \begin{bmatrix} \mathbf{p}^T & \mathbf{v}^T & \mathbf{q}^T & \mathbf{b}_a^T & \mathbf{b}_g^T \end{bmatrix}^T
\end{equation}
where $\mathbf{p}, \mathbf{v} \in \mathbb{R}^3$ denote the 3D position and linear velocity, $\mathbf{q}$ represents the orientation quaternion, and $\mathbf{b}_a, \mathbf{b}_g \in \mathbb{R}^3$ are the accelerometer and gyroscope biases, respectively. Accordingly, the error-state vector $\delta\mathbf{x} \in \mathbb{R}^{15}$ is parameterized using a minimal 3D representation for the orientation error $\delta\boldsymbol{\theta}$:
\begin{equation}
\delta\mathbf{x} = \begin{bmatrix} \delta\mathbf{p}^T & \delta\mathbf{v}^T & \delta\boldsymbol{\theta}^T & \delta\mathbf{b}_a^T & \delta\mathbf{b}_g^T \end{bmatrix}^T.
\end{equation}

During the prediction step, high-frequency IMU measurements are integrated to propagate the nominal state and its covariance. This prediction serves as the initial guess for the point cloud alignment process and allows for the dynamic deskewing of the incoming LiDAR scans. Once a new scan is registered against the G-EDF map, the resulting 6-DoF optimized pose (position and rotation) is fed back into the ESKF for the update step.

\subsection{Point Cloud Registration and Optimization}
\label{sec:optimization}

The alignment of the incoming point cloud against the global map is formulated as a direct registration problem. Instead of relying on feature extraction or explicit point-to-point correspondences, the system directly evaluates the distance field $\hat{d}(\mathbf{x})$ parameterized by the G-EDF representation.

Let $\mathcal{P} = \{\mathbf{p}_1, \mathbf{p}_2, \dots, \mathbf{p}_P\}$ be the preprocessed, deskewed point cloud. The objective is to find the optimal 6-DoF rigid transformation $\mathbf{T}^* = (\mathbf{t}^*, \mathbf{q}^*)$, comprising a translation vector $\mathbf{t} \in \mathbb{R}^3$ and an orientation quaternion $\mathbf{q}$, that minimizes the squared distance of the transformed points to the map's implicit surface. This is achieved by solving the following non-linear least squares problem:
\begin{equation}
\mathbf{T}^* = \arg \min_{\mathbf{t}, \mathbf{q}} \sum_{i=1}^{P} \rho \left( \hat{d}\big(\mathbf{R}(\mathbf{q})\mathbf{p}_i + \mathbf{t}\big)^2 \right)
\end{equation}
where $\mathbf{R}(\mathbf{q})\mathbf{p}_i + \mathbf{t}$ denotes the transformation of the local point $\mathbf{p}_i$ into the global coordinate frame, $\hat{d}(\cdot)$ is the continuous distance queried from the G-EDF map, and $\rho(\cdot)$ is a robust loss function designed to mitigate the influence of outliers.

To ensure robust convergence, even under poor initial guesses or in the presence of unmapped dynamic objects, the optimization leverages a two-stage coarse-to-fine strategy underpinned by a continuous dynamic outlier rejection mechanism:

\begin{itemize}
    \item \textbf{Continuous Dynamic Outlier Masking:} Throughout both optimization stages, points that fall outside the mapped volume are assigned a zero gradient, effectively ignoring them. However, they are not permanently discarded; if subsequent alignment adjustments bring them back into the valid map domain, they seamlessly contribute to the optimization again.
    \item \textbf{Coarse Stage:} The optimization is initialized with a wide Cauchy scale parameter for the loss function $\rho(\cdot)$. This relaxed configuration expands the basin of attraction, enabling the optimizer to successfully capture significant rotational or translational offsets (e.g., when IMU data is unavailable) without being trapped in local minima.
    \item \textbf{Fine Stage:} Once the coarse alignment converges, a stricter Cauchy scale is applied. This phase aggressively penalizes remaining within-map outliers (such as sensor noise or unmapped dynamic structures) and fine-tunes the transformation to achieve high-precision registration.
\end{itemize}

By exploiting the exact analytical gradients $\nabla \hat{d}(\mathbf{x})$ provided by the G-EDF formulation, the Ceres-based solver computes the required Jacobians efficiently, ensuring rapid and stable convergence across both stages.

\section{Experimental Results}
\label{sec:experiments}

This section evaluates the proposed G-EDF-Loc framework in terms of distance field fidelity, gradient consistency, and localization performance. The framework is implemented in C++17 and optimized for multi-core CPUs using OpenMP. All experiments were performed on an HP Victus 16 laptop equipped with 32 GB RAM and a 13th-generation Intel Core i7-13700H processor. No GPU was used for the experimentation, demonstrating the computational efficiency of the approach.

\subsection{Datasets}
\label{sec:datasets}

The evaluation of the proposed G-EDF representation and its subsequent application in the localization system was conducted using the Newer College \cite{zhang2021multicamera} and Snail datasets \cite{huai2025snailradarlargescalediverse}. For the Newer College dataset, the ``New College, Oxford'' campus was mapped covering an approximate area of $315 \times 260 \times 40$ m$^3$. Localization performance was assessed using the \textit{quad-medium}, \textit{quad-hard}, and \textit{park} sequences. The \textit{quad-medium} sequence follows a standard trajectory with a series of loops in a controlled environment; \textit{quad-hard} evaluates the system under more hostile conditions, featuring fast walking with aggressive motions and close wall approaches; finally, \textit{park} consists of a long experiment spanning the entire park and two quads with multiple loops.

Regarding the Snail dataset, the mapped area is approximately $900 \times 605 \times 120$ m$^3$, and the evaluated trajectories include \textit{bc} (20231007/4), \textit{sl} (20231007/2), \textit{ss} (20231109/4), and \textit{st} (20240113/1). Both datasets feature highly heterogeneous maps, comprising confined environments surrounded by tall buildings as well as open spaces with dense vegetation. 

\subsection{Reconstruction Accuracy}
\label{sec:reconstruction_accuracy}

The fidelity of the continuous approximation is evaluated against the dense ground truth point cloud. The G-EDF map is queried at uniformly sampled positions ($0.3$ m step) throughout the volume, comparing the inferred distance $\hat{d}(\mathbf{x})$ against the nearest neighbor distance in the ground truth mesh. To ensure robust statistics, the top $0.01\%$ of outliers are excluded. Table \ref{tab:metrics} summarizes the error metrics.

For the New College dataset, a Mean Absolute Error (MAE) of $0.033$ m is achieved. Notably, the median error is significantly lower at $0.018$ m, indicating that the vast majority of the environment is reconstructed with centimeter-level precision. Consistent performance is observed in the large-scale Snail dataset, yielding an MAE of $0.035$ m and a similar median error of $0.018$ m. This demonstrates the framework's scalability, maintaining high fidelity regardless of the environment size.

To ensure the field's suitability for optimization, the quality of the derivative is validated by computing the mean gradient magnitude $\|\nabla \hat{d}\|$ over the entire domain. A perfect distance field must satisfy the Eikonal equation $\|\nabla d\| = 1$. As shown in Table \ref{tab:metrics}, the mean gradient magnitude remains close to unity ($0.984$ for ``New College'' and $0.979$ for ``Snail''), confirming that G-EDF provides a mathematically consistent field with stable gradients, a prerequisite for robust gradient-based trajectory optimization.

\begin{table}[t]
\caption{Quantitative results on reconstruction fidelity. Distance errors are in meters ($\downarrow$ lower is better). Gradient magnitude ($\|\nabla \hat{d}\|$) indicates Eikonal consistency ($\approx 1.0$ is ideal).}
\label{tab:metrics}
\centering
\small
\setlength{\tabcolsep}{3pt}
\begin{tabular}{@{}lccccc@{}}
\toprule
\multirow{2}{*}{\textbf{Dataset}} & \multicolumn{3}{c}{\textbf{Distance Error} (m)} & \multicolumn{2}{c}{\textbf{Gradient} $\|\nabla \hat{d}\|$} \\
\cmidrule(lr){2-4} \cmidrule(l){5-6}
 & MAE $\downarrow$ & Med. $\downarrow$ & Std $\downarrow$ & Mean & Std $\downarrow$ \\
\midrule
New College & 0.033 & 0.018 & 0.044 & 0.984 & 0.089 \\
Snail       & 0.035 & 0.018 & 0.049 & 0.979 & 0.108 \\
\bottomrule
\end{tabular}
\end{table}

\subsection{Localization Evaluation}
\label{sec:localization_evaluation}

\begin{figure*}[t!]
    \centering
    \begin{subfigure}[b]{0.24\textwidth}
        \centering
        \includegraphics[width=\textwidth]{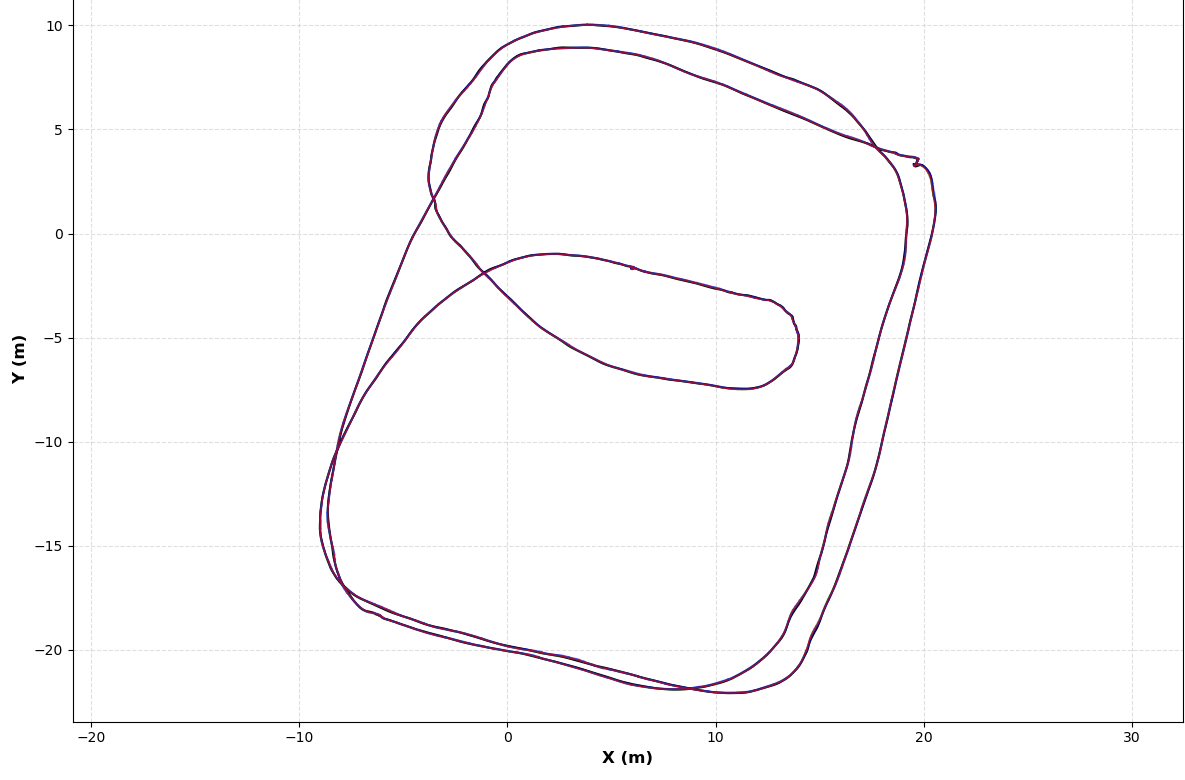}
        \caption{\textit{bc} - Inertial}
        \label{fig:traj_bc_inertial}
    \end{subfigure}
    \hfill
    \begin{subfigure}[b]{0.24\textwidth}
        \centering
        \includegraphics[width=\textwidth]{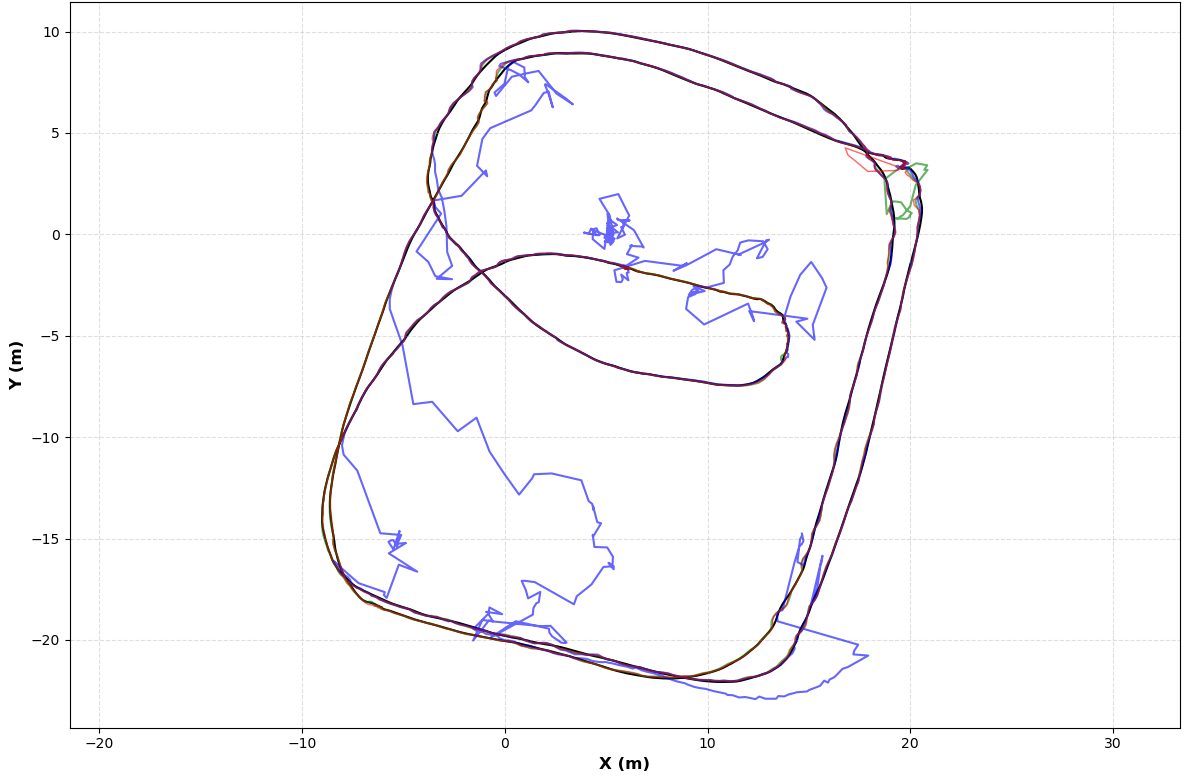}
        \caption{\textit{bc} - No IMU}
        \label{fig:traj_bc_noimu}
    \end{subfigure}
    \hfill
    \begin{subfigure}[b]{0.24\textwidth}
        \centering
        \includegraphics[width=\textwidth]{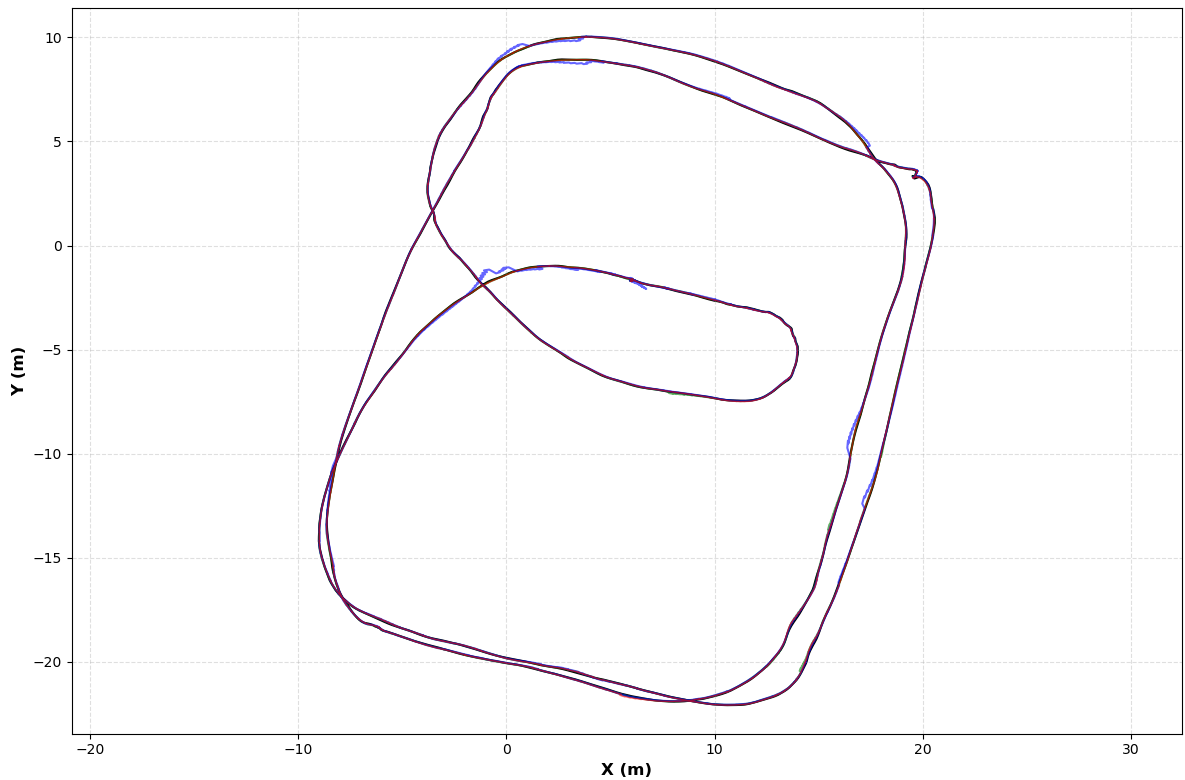}
        \caption{\textit{bc} - Low Noise}
        \label{fig:traj_bc_low}
    \end{subfigure}
    \hfill
    \begin{subfigure}[b]{0.24\textwidth}
        \centering
        \includegraphics[width=\textwidth]{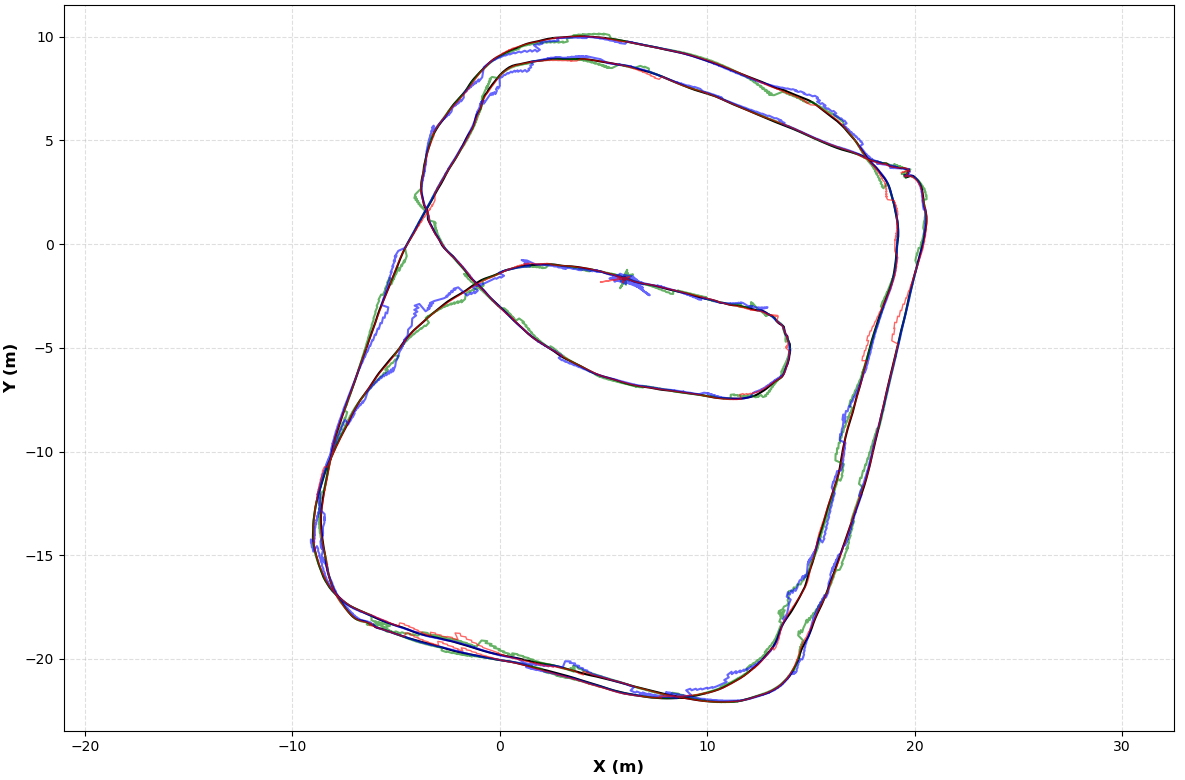}
        \caption{\textit{bc} - High Noise}
        \label{fig:traj_bc_high}
    \end{subfigure}

    \vspace{0.2cm} 

    \begin{subfigure}[b]{0.24\textwidth}
        \centering
        \includegraphics[width=\textwidth]{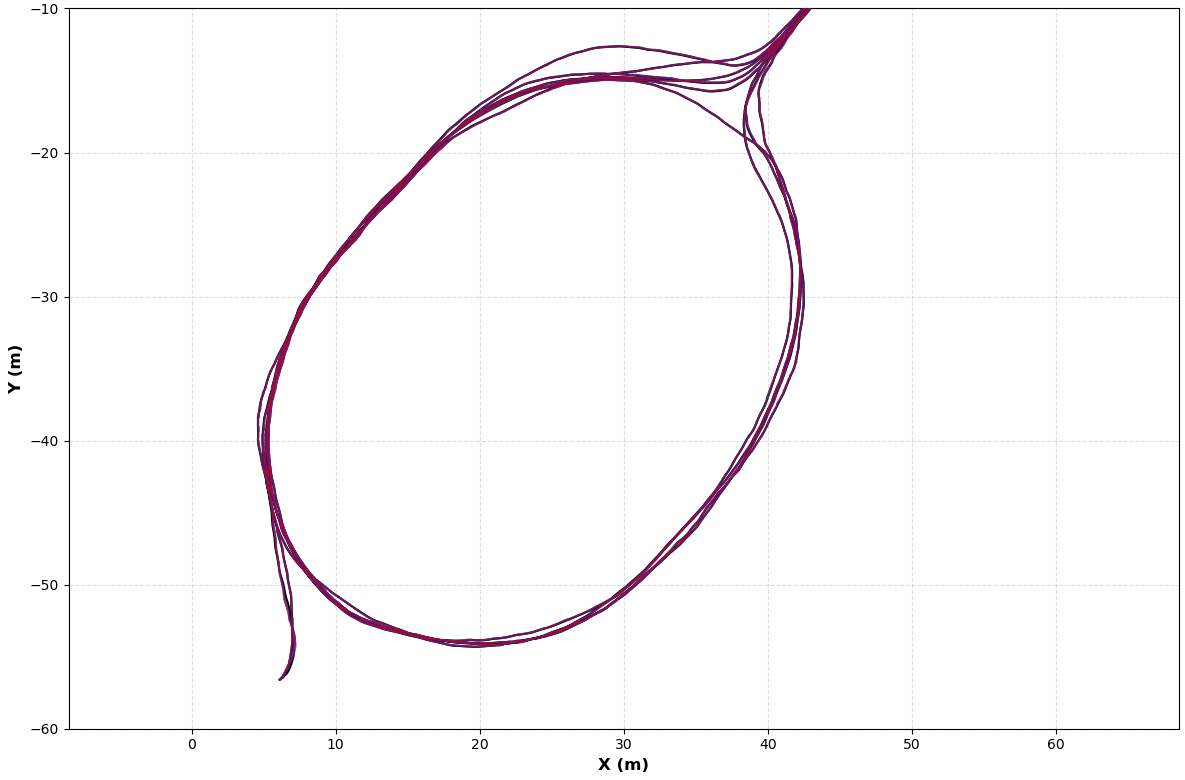}
        \caption{\textit{park} - Inertial}
        \label{fig:traj_park_inertial}
    \end{subfigure}
    \hfill
    \begin{subfigure}[b]{0.24\textwidth}
        \centering
        \includegraphics[width=\textwidth]{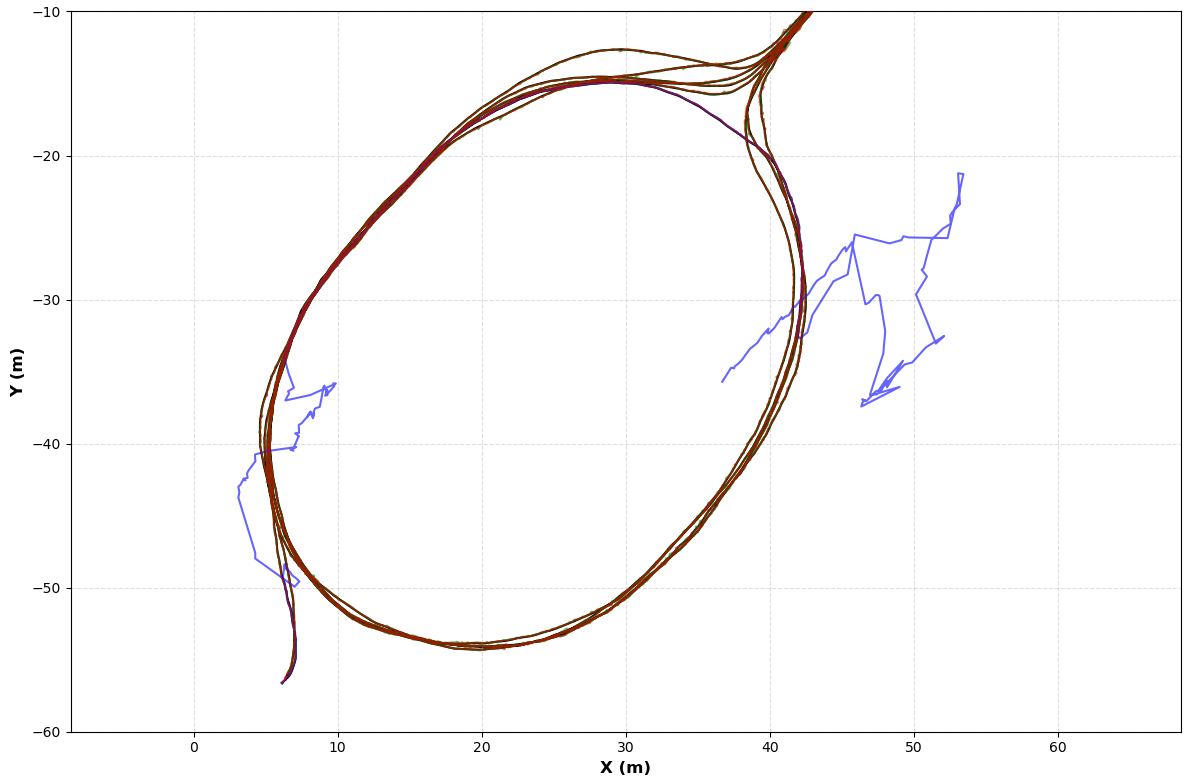}
        \caption{\textit{park} - No IMU}
        \label{fig:traj_park_noimu}
    \end{subfigure}
    \hfill
    \begin{subfigure}[b]{0.24\textwidth}
        \centering
        \includegraphics[width=\textwidth]{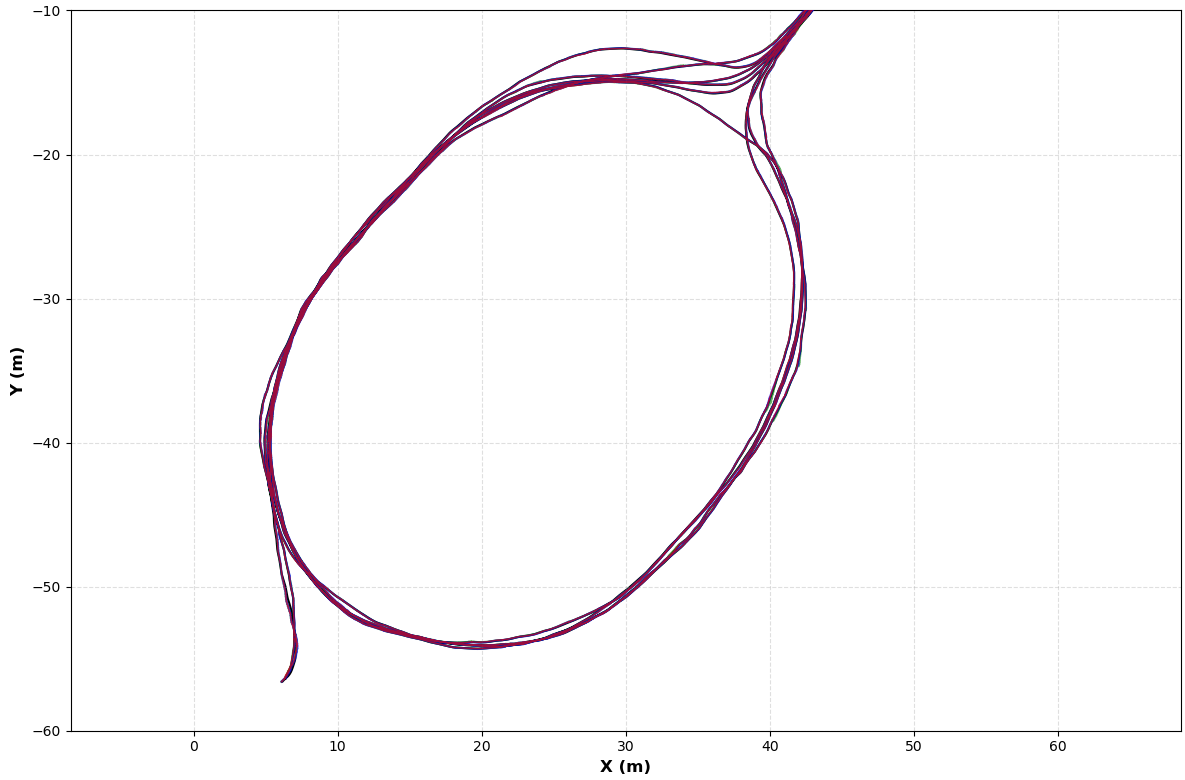}
        \caption{\textit{park} - Low Noise}
        \label{fig:traj_park_low}
    \end{subfigure}
    \hfill
    \begin{subfigure}[b]{0.24\textwidth}
        \centering
        \includegraphics[width=\textwidth]{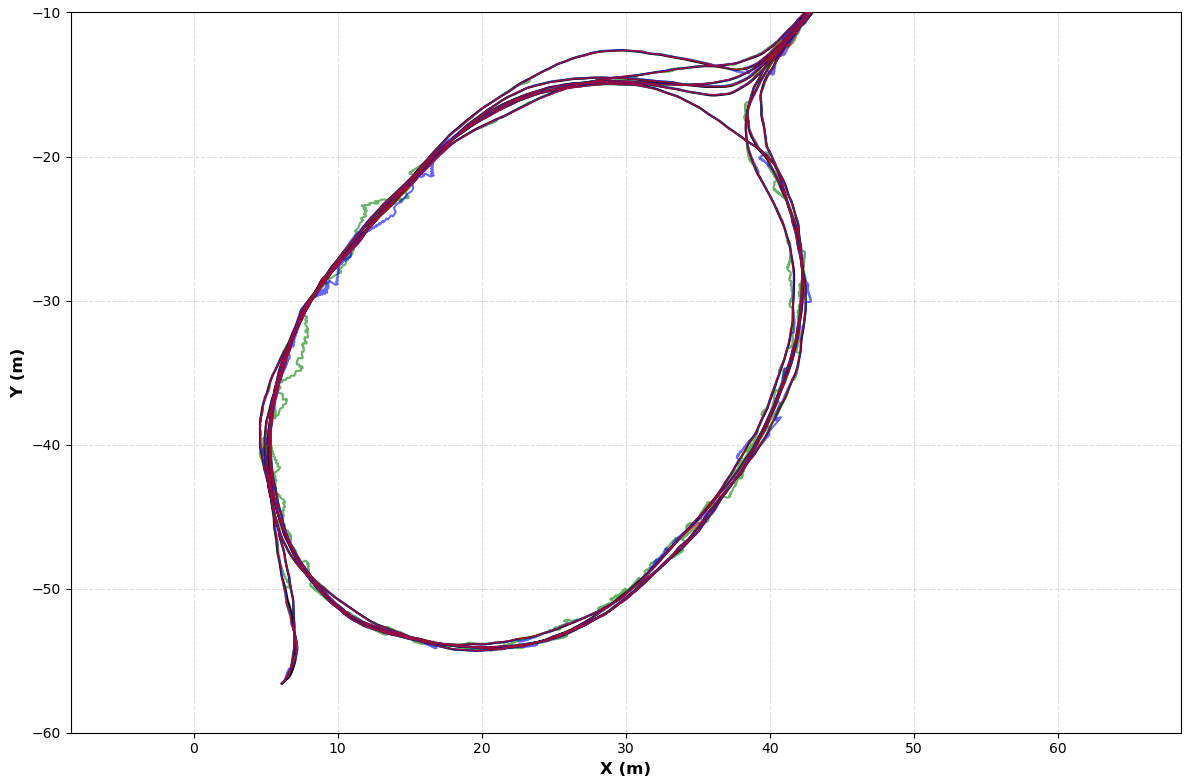}
        \caption{\textit{park} - High Noise}
        \label{fig:traj_park_high}
    \end{subfigure}
    
    \caption{Qualitative 2D trajectory comparison featuring the full \textit{bc} sequence (top row) and a zoomed-in detail of the \textit{park} sequence (bottom row). Trajectories are depicted as follows: ground truth (black), the proposed G-EDF-Loc (red), Fast-GICP (blue), and NDT (green).}
    \label{fig:trajectory_comparison}
\end{figure*}

Having evaluated the reconstruction accuracy of the G-EDF representation, this section assesses the performance of the proposed localization pipeline. To benchmark our approach against established state-of-the-art registration algorithms, specifically Fast-GICP and the Normal Distributions Transform (NDT), the efficient implementations provided by the widely adopted \textit{hdl\_localization} package \cite{koide2019portable} were used.

To ensure a strictly fair comparison and isolate the effect of the optimization algorithms, the localization wrapper is identical for all evaluated methods. While we integrate the core NDT and Fast-GICP registration modules, the tracking pipeline, including the Error-State Kalman Filter (ESKF) for prior pose prediction and the point cloud preprocessing steps, such as deskewing and downsampling (which explicitly dictates the quantity of points the model works with), remains exactly the same as in our G-EDF-Loc pipeline. This unified setup allows us to accurately evaluate the algorithms under identical sensor configurations, both in nominal conditions and through a comprehensive robustness study designed to assess resilience against degraded or completely absent initial IMU odometry priors.

Accordingly, the three approaches were evaluated across the trajectories detailed in Section \ref{sec:datasets}. The following experimental setups were defined to assess their performance under varying conditions:

\begin{itemize}
    \item \textbf{Standard Setup:} This configuration evaluates the algorithms under nominal conditions. It represents the ideal operation of the unified tracking pipeline, where the Error-State Kalman Filter (ESKF) provides a robust initial pose prediction and enables the dynamic deskewing of the incoming LiDAR scans prior to the execution of the respective registration module (G-EDF-Loc, Fast-GICP, or NDT).
    
    \item \textbf{NoOdom:} In this setup, the IMU is completely omitted, meaning the ESKF is not utilized for odometry estimation or dynamic deskewing. Instead, the tracking is initialized at the correct position, and the system relies solely on the optimized pose from the previous scan as the initial guess for the current point cloud registration.
    
    \item \textbf{Gaussian Noise:} To simulate degraded odometry, artificial Gaussian noise is injected into the position and yaw angle of the initial guess estimated by the ESKF. We evaluated two distinct degradation levels: a \textit{Low Noise} profile with a standard deviation of $0.25$ m in each translation axis and $0.05$ rad in yaw, and a \textit{High Noise} profile with a standard deviation of $0.5$ m and $0.1$ rad, respectively.
\end{itemize}

Across all experiments, the performance is measured using the Root Mean Square Error (RMSE) for position (in meters) and rotation (in degrees), alongside the average computational time required to process each scan. 

For the map representation, our approach utilizes the continuous G-EDF, which provides explicit distance information and global $C^{1}$ continuity within the modeled bounds. In contrast, the baseline methods rely on a discrete global point cloud map. Because Fast-GICP was unable to process the raw, highly dense map (comprising approximately $141$ million points), and NDT suffered from severe computational overhead without any noticeable gain in position or rotation accuracy, the point cloud map was downsampled using a $0.1$ m voxel grid. During the sequential scan-to-map alignment for localization, we standardize the incoming LiDAR scans using a $0.5$ m downsampling resolution for all approaches. Adjusting this resolution directly dictates the quantity of points the approach works with during each registration step. While we initially evaluated our G-EDF-Loc approach using a denser $0.1$ m scan resolution, the localization metrics were practically identical but with higher computational cost. Notably, even under hostile conditions with missing IMU data or highly noisy priors, where a denser point cloud typically aids in correcting erroneous initial guesses, our continuous optimization algorithm converged robustly using only the $0.5$ m downsampled scans.

To ensure a fair and reproducible benchmark, the baseline algorithms from the \textit{hdl\_localization} package were primarily configured using their default parameters. For the NDT algorithm, the resolution was set to $1.0$ m, with a step size of $0.1$, and a transformation epsilon of 1e-3. Additionally, the \texttt{DIRECT7} neighborhood search method was employed, as it is highly optimized for downsampled maps. Similarly, Fast-GICP was configured with a transformation epsilon of 1e-3. Both methods heavily utilized multi-threading to maximize their computational efficiency.
\begin{table*}[t]
\centering
\caption{Robustness Study: Evaluation of G-EDF-Loc against Fast-GICP and NDT across trajectories (0.5 m resolution). Systems are tested under normal operation (Inertial), missing IMU, and degraded priors (Low/High Gaussian noise). Best and second-best results are \textbf{bold} and \underline{underlined} [Pos: m, Rot: °, Time: ms].}\label{tab:robustness_study}
\footnotesize
\setlength{\tabcolsep}{3.5pt} 
\begin{tabular}{@{}ll ccc ccc ccc ccc@{}}
\toprule
\multirow{2}{*}{\textbf{Traj.}} & \multirow{2}{*}{\textbf{Method}} & \multicolumn{3}{c}{\textbf{Inertial} $\downarrow$} & \multicolumn{3}{c}{\textbf{No IMU} $\downarrow$} & \multicolumn{3}{c}{\textbf{Low} $\downarrow$} & \multicolumn{3}{c}{\textbf{High} $\downarrow$} \\
\cmidrule(lr){3-5} \cmidrule(lr){6-8} \cmidrule(lr){9-11} \cmidrule(l){12-14}
 &                     & Pos            & Rot            & Time          & Pos            & Rot & Time & Pos & Rot & Time & Pos & Rot & Time \\ 
\midrule

\multirow{3}{*}{bc} 
 & Ours  & \underline{0.084} & \underline{1.133} & \underline{76.9}  & \textbf{0.215} & \textbf{1.204} & \underline{95.2}  & \textbf{0.084} & \textbf{1.133} & 107.3          & \textbf{0.118} & \textbf{1.207} & \textbf{118.5} \\
 & Fast-GICP      & \textbf{0.082}    & \textbf{1.129}    & \textbf{67.7}     & 2.884          & 38.900         & 1366.9         & 0.122          & 1.179          & \textbf{78.7}  & 0.261          & 1.452          & \underline{134.6} \\
 & NDT            & \underline{0.084} & 1.138             & 94.7              & \underline{0.308} & \underline{3.349} & \textbf{81.5}  & \underline{0.087} & \underline{1.134} & \underline{98.9}  & \underline{0.169} & \underline{1.241} & 134.7 \\ 
\midrule

\multirow{3}{*}{sl} 
 & Ours & \underline{0.157} & 1.049          & \textbf{61.8}  & \textbf{0.217} & \textbf{0.985} & \textbf{63.3}  & \textbf{0.158} & \underline{1.048} & \textbf{70.6}  & \textbf{0.192} & \textbf{1.089} & \textbf{83.9} \\
 & Fast-GICP      & \textbf{0.153} & \textbf{1.043} & \underline{63.5}  & -              & -              & -              & 0.169          & \textbf{1.045} & \underline{75.2}  & 0.320          & 1.306          & 148.9 \\
 & NDT           & 0.164          & \underline{1.044} & 76.7           & -              & -              & -              & \underline{0.165} & \textbf{1.045} & 83.1           & \underline{0.219} & \underline{1.167} & \underline{103.3} \\ 
\midrule

\multirow{3}{*}{st} 
  & Ours & 0.140 & \underline{0.500}    & \textbf{57.9}     & \textbf{0.147}    & \underline{0.393}    & \textbf{61.4}     & \textbf{0.140} & \textbf{0.500}    & \textbf{72.6}     & \textbf{0.165}    & \underline{0.774}    & \textbf{86.0} \\
 & Fast-GICP      & \textbf{0.137} & \textbf{0.496} & \underline{61.2}  & -              & -              & -              & \underline{0.141}          & \underline{0.513} & \underline{85.2}  & 0.319          & \textbf{0.744} & 262.5 \\
 & NDT            & \underline{0.139}         & \underline{0.500} & 85.4           & \underline{0.154} & \textbf{0.369} & \underline{73.8}  & \textbf{0.140} & \textbf{0.500} & 93.6           & \underline{0.210} & 1.011 & \underline{112.2} \\ 
\midrule

\multirow{3}{*}{ss} 
 & Ours & \underline{0.114} & 1.032          & \textbf{66.7}  & \textbf{0.176} & \textbf{1.677} & \textbf{64.4}  & \underline{0.114} & \underline{1.029 }         & \textbf{85.8}  & \textbf{0.144} & \textbf{1.110 }         & \textbf{103.6} \\

 & Fast-GICP     & \textbf{0.101}    & \textbf{1.015} & 71.0              & -                 & -                 & -                 & 0.177             & 1.077             & \underline{89.1}  & -                 & -                 & - \\
 & NDT            & \textbf{0.101}    & \underline{1.018}             & \underline{70.1}  & -                 & -                 & -                 & \textbf{0.105}    & \textbf{1.026} & 90.1              & \underline{0.198} & \underline{1.176} & \underline{126.6} \\ 
\midrule

\multirow{3}{*}{quad-medium} 
 & Ours & 0.111             & \textbf{1.950}    & \underline{57.5}  & \underline{0.118} & \textbf{2.195}    & \textbf{56.4}     & \underline{0.107} & \textbf{1.887}    & \underline{58.4}  & \textbf{0.107}    & \textbf{1.890}    & \textbf{57.8} \\
 & Fast-GICP      & \underline{0.110} & 1.965             & \textbf{44.5}     & -                 & -                 & -                 & 0.111             & 1.965             & \textbf{49.7}     & \underline{0.117} & \underline{1.993} & \underline{66.5} \\
 & NDT           & \textbf{0.104}    & \underline{1.961} & 65.7              & \textbf{0.115}    & \underline{2.203} & \underline{59.7}  & \textbf{0.104}    & \underline{1.964} & 71.7              & 0.187             & 2.221             & 86.9 \\ 
\midrule

\multirow{3}{*}{quad-hard} 
 & Ours                 & \textbf{0.105}    & \textbf{2.630}    & \textbf{40.8}     & \textbf{0.131}    & \textbf{3.142}    & \textbf{37.6}     & \textbf{0.106}    & \textbf{2.628}    & \textbf{44.6}     & \textbf{0.106}    & \textbf{2.620}    & \textbf{46.5} \\
 & Fast-GICP            & 0.114             & \underline{2.805} & 73.7              & -                 & -                 & -                 & 0.116             & \underline{2.797} & \underline{67.5}  & 0.206             & \underline{2.767} & 81.9 \\
 & NDT                  & \underline{0.110} & 2.847             & \underline{62.1}  & -                 & -                 & -                 & \underline{0.112} & 2.823             & 68.1              & \underline{0.180} & 2.832             & \underline{81.4} \\ 
\midrule

\multirow{3}{*}{park} 
 & Ours        & \underline{0.099} & \underline{1.321} & \textbf{46.3}     & \textbf{0.119}    & \textbf{1.408}    & \textbf{52.7}     & \underline{0.099} & \underline{1.289} & \textbf{48.6}     & \textbf{0.099}    & \textbf{1.290}    & \textbf{51.7} \\
 & Fast-GICP            & 0.117             & 1.406             & \underline{59.7}  & -                 & -                 & -                 & 0.113             & 1.311             & 68.1              & 0.195             & \underline{2.423} & \underline{82.1} \\
 & NDT                  & \textbf{0.094}    & \textbf{1.288}    & 62.7              & \underline{0.139} & \underline{1.586} & \underline{61.9}  & \textbf{0.094}    & \textbf{1.288}    & \underline{62.7}  & \underline{0.142} & 2.481             & 94.4 \\ 
\bottomrule
\end{tabular}
\end{table*}

The quantitative results detailed in Table \ref{tab:robustness_study} demonstrate the performance of the evaluated algorithms across these varying levels of odometry degradation. Under nominal conditions (Standard Setup), all three approaches perform competitively, achieving State-of-the-Art (SoA) localization accuracy with translation and rotation errors varying only at the centimeter level. In this ideal scenario, both G-EDF-Loc and Fast-GICP demonstrate significantly faster processing times compared to NDT.

However, the resilience of the proposed approach becomes distinctly evident under suboptimal odometry. While all algorithms maintain track under the \textit{Low Noise} profile, the baseline methods begin to exhibit increased computational costs. This degradation is severely amplified under \textit{High Noise}, where Fast-GICP and NDT struggle to align correctly, resulting in unstable pose estimates characterized by visible spikes and sharp jumps, unlike the smooth trajectory maintained by our proposed method. This erratic behavior drives the baselines toward divergence in certain situations. The most demanding evaluation, the \textbf{NoOdom} setup, further highlights these limitations. Although the baselines can converge in stable, feature-dense scenarios lacking sharp turns, they frequently fail under aggressive motions, as seen in the \textit{sl}, \textit{st}, and \textit{quad-hard} trajectories. In contrast, G-EDF-Loc proves highly resilient, successfully converging across varied trajectories without any initial prior. Crucially, even during wide-baseline alignments, while the processing time of our continuous optimization algorithm does experience a moderate and bounded increase, it successfully avoids the severe computational bottlenecks exhibited by the baselines. This stable behavior ensures the system preserves its real-time capabilities while maintaining SoA precision. To better visualize this overall behavior and the performance disparities across the different setups, a qualitative 2D trajectory comparison is presented in Fig. \ref{fig:trajectory_comparison}, illustrating the estimated poses for the \textit{bc} sequence from the Snail dataset and a representative segment of the \textit{park} sequence from the Newer College dataset.


\section{Conclusions}
\label{sec:conclusions}

This paper presented G-EDF-Loc, a robust 6-DoF localization framework built upon a novel continuous 3D distance field representation. By modeling the environment through a Block-Sparse Gaussian Mixture Model with axis-aligned covariances, the approach achieves a memory-efficient Euclidean Distance Field that preserves $C^1$ continuity across transition regions. The proposed formulation provides analytical gradients, effectively mitigating the discretization artifacts and local minima that typically hinder standard grid-based methods.

Experimental evaluations across large-scale, heterogeneous datasets demonstrated that the G-EDF representation yields centimeter-level distance field fidelity while empirically maintaining Eikonal consistency. Furthermore, when integrated into a direct scan-to-map registration pipeline, the system exhibited exceptional resilience. While state-of-the-art discrete baselines, such as Fast-GICP and NDT, suffered from severe computational bottlenecks or divergent pose estimates under degraded odometry, G-EDF-Loc consistently maintained accurate, real-time tracking. The continuous optimization algorithm proved highly robust, successfully converging even under high-noise conditions and in the complete absence of IMU priors.

Future work will focus on further optimizing the block-sparse hashing mechanism to handle even larger environments with minimal memory overhead. Additionally, the exploration of multi-resolution Gaussian representations could provide even faster convergence and increase the basin of attraction for initial pose estimation.

\balance


\bibliographystyle{IEEEtran}
\bibliography{bibliography}

\end{document}